\documentclass[10pt,twocolumn,letterpaper]{article}

\usepackage{iccv}              
\usepackage{multirow}
\usepackage{booktabs} 
\usepackage{caption}
\definecolor{darkgreen}{rgb}{0,0.5,0}
\usepackage{amsmath}
\usepackage{amssymb}
\usepackage{algorithm}

\usepackage[accsupp]{axessibility}
\definecolor{promptcolor}{RGB}{200, 235, 227}
\definecolor{promptcolor2}{RGB}{187,227,252}
\definecolor{prompttitlecolor}{RGB}{175, 172, 172}
\usepackage[framemethod=TikZ]{mdframed} 

\usepackage{algorithmicx}
\usepackage{algpseudocode}
\newcommand{\method}{\textsc{TIR}}
\newcommand{\methodwithspace}{\textsc{TIR }}
\definecolor{iccvblue}{rgb}{0.21,0.49,0.74}
\usepackage[pagebackref,breaklinks,colorlinks,allcolors=iccvblue]{hyperref}

\def\paperID{7} 
\def\confName{ICCV}
\def\confYear{2025}

\mdfdefinestyle{prompt}{%
    linecolor =   black,
    linewidth =   0.2pt,
    backgroundcolor =  promptcolor2,
    roundcorner     =   1pt,
    font = \small\ttfamily,
    innerleftmargin=1.5pt,
    innerrightmargin=1.5pt
}

\newmdenv[%
    roundcorner=5pt, 
    linecolor =   black,
    linewidth =   1pt,
    font = \small\ttfamily,
    subtitlebackgroundcolor=prompttitlecolor, 
    frametitlebackgroundcolor=prompttitlecolor,
    backgroundcolor=promptcolor, 
    frametitle={Generated caption},
    subtitleaboveskip=0.5\baselineskip,
    subtitlebelowskip=0.5\baselineskip,
    ]{captionenv}

\DeclareMathAlphabet{\pazocal}{OMS}{zplm}{m}{n}

\DeclareMathAlphabet\mathbfcal{OMS}{cmsy}{b}{n}

\title{Test-time Prompt Refinement for Text-to-Image Models}

\author{%
  Mohammad Abdul Hafeez Khan$^{1}$\footnotemark[1]%
  \quad Yash Jain$^{2}$\footnotemark[1]%
  \quad Siddhartha Bhattacharyya$^{1}$%
  \quad Vibhav Vineet$^{2}$\\
  $^{1}$Florida Institute of Technology, USA\\
  $^{2}$Microsoft Research, USA
}

\begin{document}


\twocolumn[{
	\maketitle
	\vspace{-2em}
	\renewcommand\twocolumn[1][]{#1}
    \begin{center}
    \centering
    \includegraphics[width=\textwidth]{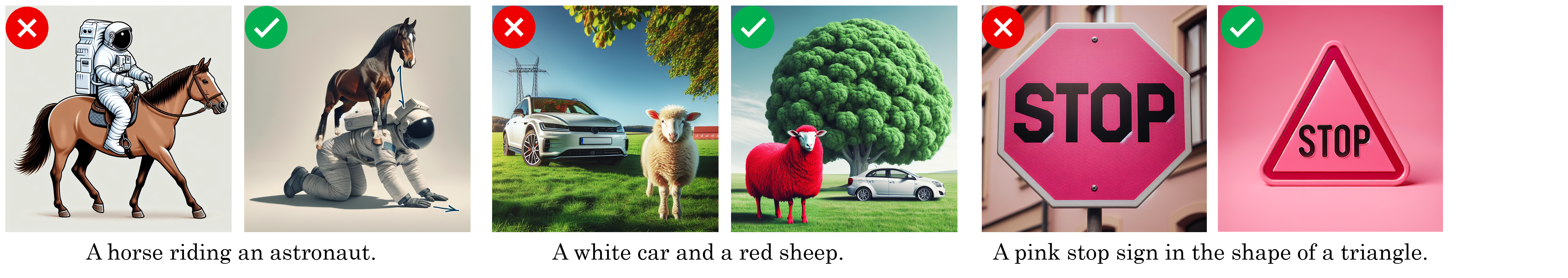}
    \vspace{-10pt}
    \captionof{figure}{\textbf{Effectiveness of the proposed Test-time Image Refinement (TIR) approach in enhancing image generation quality.} TIR iteratively refines input prompts based on the generated images to more accurately reflect the given text instructions. For each prompt, the left image shows the original DALL-E 3 output, while the right image presents the refined result produced by our approach (TIR), demonstrating improved content control and better adherence to the specified prompt. Best viewed in color.}
    \label{fig:intro}
\end{center}
\vspace{-0.1cm}

}]
\footnotetext[1]{* Equal contribution. Corresponding author: \texttt{mkhan@my.fit.edu}}

\begin{abstract}
\vspace{-0.15em}
Text-to-image (T2I) generation models have made significant strides but still struggle with prompt sensitivity: even minor changes in prompt wording can yield inconsistent or inaccurate outputs. To address this challenge, we introduce a closed-loop, test-time prompt refinement framework that requires no additional training of the underlying T2I model, termed \method. In our approach, each generation step is followed by a refinement step, where a pretrained multimodal large language model (MLLM) analyzes the output image and the user’s prompt. The MLLM detects misalignments (e.g., missing objects, incorrect attributes) and produces a refined and physically grounded prompt for the next round of image generation.
By iteratively refining the prompt and verifying alignment between the prompt and the image, \methodwithspace corrects errors, mirroring the iterative refinement process of human artists.
We demonstrate that this closed-loop strategy improves alignment and visual coherence across multiple benchmark datasets, all while maintaining plug-and-play integration with black-box T2I models. Code is available at \url{https://github.com/hafeezkhan909/Test-time-Image-Refinement}.
\end{abstract}   
\vspace{-0.3cm}
\section{Introduction}  
\label{sec:intro}  

Text-to-image (T2I) models have enabled users to create high-quality images from textual descriptions \cite{rombach2022high, ramesh2022hierarchical, ramesh2021zero, reed2016generative, flux2024}. However, these models often fail when prompts require compositional reasoning or commonsense understanding. For instance, we observe that given the prompt \textit{“a horse riding an astronaut”}, a T2I model often generates an astronaut riding a horse instead, as shown in Fig.~\ref{fig:intro}, thereby reversing the intended subject-object relationship. This occurs because the model defaults to more statistically frequent or visually plausible configurations learned during training, rather than faithfully interpreting the prompt's semantic structure. Such errors reflect a broader inability to interpret uncommon spatial relations, especially when they conflict with learned priors or require understanding of syntactic cues, causal direction, or atypical physical dynamics.

This reveals a core limitation of current T2I systems: they operate in an open-loop fashion and lack the ability to assess prompt plausibility or self-correct errors after generation. Even advanced models like DALL-E 3 \cite{dalle3}, Stable Diffusion \cite{rombach2022high}, and Flux \cite{flux2024} often produce images that are photorealistic yet semantically incorrect. This is especially evident in prompts involving spatial relationships, counting, negation, or temporal logic, where adherence to the prompt breaks down despite generating high-quality images.

These challenges indicate that T2I models struggle with the implicit semantics of prompts unless such constraints are made explicit, which better aligns with training-time prompt-image supervision based on direct token-to-visual feature associations \cite{radford2021learning, ramesh2022hierarchical, rombach2022high}. Existing solutions such as fine-tuning \cite{lee2023aligning, fan2023dpok}, post-hoc editing \cite{lian2023llm, wu2024self, avrahami2022blended, hertz2022prompt}, and prompt engineering offer partial solutions but come with trade-offs. Training-based methods improve alignment but require costly, model-specific adaptation; post-hoc corrections often distort scene structure by editing images after generation; and prompt engineering shifts the burden onto users to craft highly-specific, exhaustive prompts.

A fundamental challenge remains: \textit{Can we enable T2I models to self-correct misinterpretations iteratively, without additional training?}

\begin{figure}[!t]
    \centering
    \includegraphics[width=\linewidth]{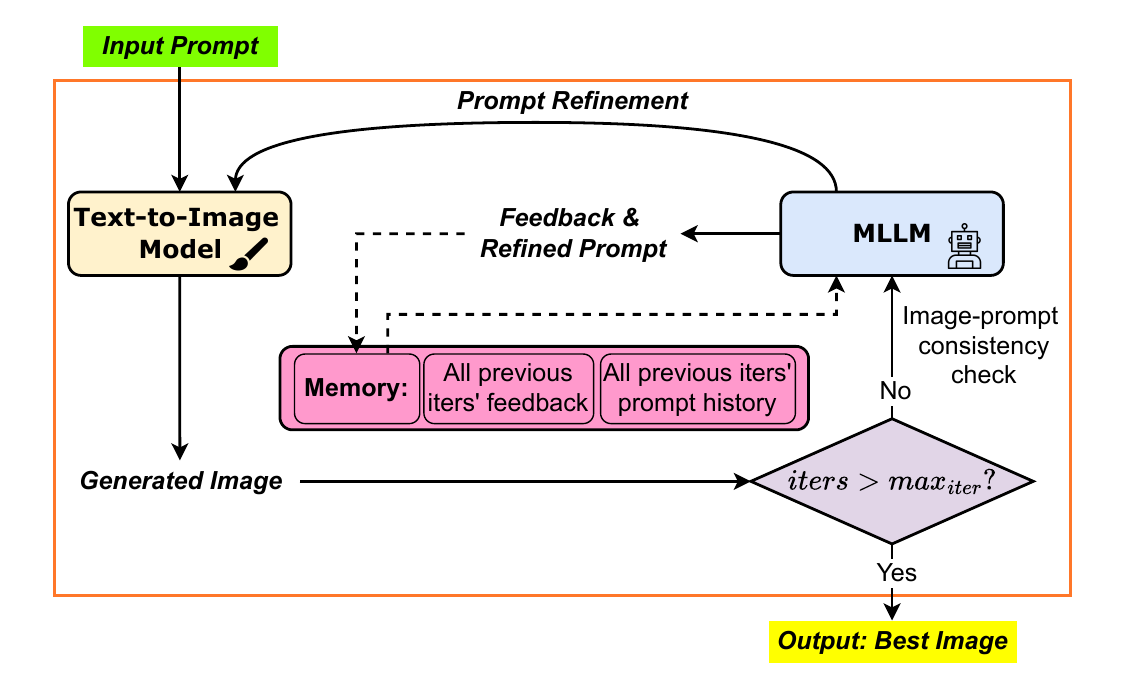}
    \caption{Test-time prompt refinement for Text-to-Image Models (\method). Starting from a user-provided input prompt, we generate an image using a text-to-image model. A Multi-modal Large Language Model (MLLM) evaluates the consistency between the output image and the prompt, and refines the prompt based on feedback. We retain all previous prompts and feedback in memory to ground the image generation in the input instructions and iteratively improve the output. We repeat this cycle for a fixed number of iterations, progressively refining the prompt and generating an image that best aligns with the original user intent.}
\vspace{-0.2cm}
    
    \label{fig:method}
\end{figure}

As MLLMs like Qwen~\citep{qwen2.5, qwen2} and GPT-4o~\cite{hurst2024gpt} continue to advance in multimodal reasoning, their understanding of text and images makes them well-suited to provide semantic feedback on the alignment between prompts and generated images. Leveraging this capability, we introduce Test-time Image Refinement (TIR), a closed-loop framework, illustrated in Fig.~\ref{fig:method}. Instead of treating T2I generation as a one-shot process, TIR uses an MLLM in each iteration to assess misalignments between the user’s intent and the generated image, focusing on semantic reasoning failures such as spatial errors, attribute mismatches, incorrect object counts, etc. It then produces a refined prompt that explicitly corrects these issues while preserving contextual history from prior refinements. This history-aware feedback loop enables the T2I model to progressively converge toward the intended scene, without relying on handcrafted prompts, scene layouts, or bounding-box supervision. 

TIR requires no additional training or architectural modifications. Instead, it treats \textit{test-time compute as a resource} and integrates seamlessly with off-the-shelf T2I models such as Stable Diffusion~\cite{rombach2022high}, DALL-E 3~\cite{dalle3}, and Flux~\cite{flux2024}, making it a flexible, plug-and-play solution for improving prompt adherence. Our key contributions are as follows:

\begin{itemize}      
    \item \textbf{MLLM-driven Test-time Refinement:} We introduce an iterative refinement framework to improve prompt-image alignment without modifying the T2I model.
    \item \textbf{Training-Free, Plug-and-Play Adaptability:} Our method requires no fine-tuning and is compatible with black-box APIs, making it applicable to any pretrained T2I model.  
    \item \textbf{Empirical Validation:} We validate our method across multiple datasets and T2I models, showing that TIR improves semantic fidelity, enables reasoning capabilities at test time and, in the future, it may help construct reasoning benchmarks for generative models. 
\end{itemize}  

\noindent Overall, our work bridges the gap between the generative strength of T2I models and the need for adaptive, self-correcting systems. By leveraging MLLM-guided refinement at test time, we enable models to iteratively improve semantic alignment, mimicking the way human artists refine their work through feedback.


\section{Literature Review}
\label{sec:literature}

\subsection{Text-to-Image (T2I) Models}
\label{sec:t2i_models}

Text-to-image generation has evolved from GAN-based models~\cite{reed2016generative, zhang2017stackgan, zhu2019dm, tao2022df} to transformer-based architectures~\cite{ding2021cogview, esser2021taming, ramesh2021zero} and diffusion-based models~\cite{ramesh2022hierarchical, podell2023sdxl, rombach2022high, saharia2022photorealistic}, significantly improving generation quality and text fidelity. However, T2I models still struggle with compositional reasoning and prompt adherence, failing to capture fine-grained spatial relationships, attribute details, and multi-object interactions~\cite{balaji2022ediff, nichol2021glide}.  

To improve controllability, researchers have explored structured conditioning signals such as keypoints, bounding boxes, scene graphs, and other domain-specific representations~\cite{chen2024training, li2023gligen, patashnik2023localizing, khan2024alina, yang2023paint, khan2022classification, yang2023reco, khan2024few, zhang2023adding}. While effective, these methods rely on external annotations that may not always be available. Training-free alternatives leverage pretrained LLMs for layout generation and spatial reasoning~\cite{feng2024layoutgpt, lian2023llm, qu2023layoutllm}. However, these approaches assume that once a layout is specified, the T2I model will faithfully follow it, overlooking potential deviations during generation. Furthermore, their latent pasting approach struggles with occlusions, and disrupts the diffusion denoising process, as the pasted noise does not align with the model’s learned score function. This misalignment introduces inconsistencies in image quality, making it difficult to maintain high fidelity throughout the entire generation.

\subsection{LLM-Driven Vision-Language Approaches}

\vspace{-0.2cm}
\paragraph{Vision-Language Grounding}  
LLMs have shown strong reasoning capabilities in vision-language tasks such as image captioning and VQA~\cite{li2023blip, alayrac2022flamingo}. Early approaches relied on CLIP-based embeddings~\cite{radford2021learning}, but recent methods~\cite{wu2023visual, koh2024generating} explore LLM-guided prompt generation for text-conditioned image synthesis. However, they still operate in a single-pass manner, where the LLM provides an initial prompt but does not iteratively refine it based on the generated image.

\vspace{-0.2cm}
\paragraph{LLM-Guided T2I Refinement}  
Recent works have incorporated LLMs into T2I pipelines to improve semantic fidelity and compositional accuracy~\cite{gani2023llm, lin2023videodirectorgpt, zhang2023controllable}. Some methods use LLMs for spatial reasoning~\cite{qu2023layoutllm, feng2024layoutgpt}, but these rely on static layouts that do not dynamically adjust based on the generated image. Others explore iterative feedback loops where an LLM evaluates and refines outputs post-generation~\cite{wu2024self, kwon2024zero, qin2024diffusiongpt}, but these approaches often struggle with maintaining visual coherence across multiple edits.

For instance, Li et al.~\cite{lian2023llm} propose a layout-first approach where an LLM generates a structured scene layout before synthesis. However, their method does not reassess alignment after generation, leading to potential misalignments that persist throughout the denoising process. Wu et al.~\cite{wu2024self} introduce an iterative feedback mechanism where an LLM refines images after they are fully generated, but their latent-space corrections disrupt the coherence of the denoising process, resulting in artifacts and inconsistencies.


Unlike prior LLM-guided generation methods, which rely on scene layouts, masks, or supervised conditioning during training, our framework operates at test time and requires no intermediate representations.




\section{Problem Statement}

Given an input text prompt \(\mathbf{c}\), the goal of a text-to-image (T2I) model is to generate an image \(\mathbf{x}\) that faithfully reflects the semantic intent of \(\mathbf{c}\). In practice, however, the generated image often fails to capture important aspects of the prompt, particularly when \(\mathbf{c}\) is under-specified, ambiguous, or reliant on implicit world knowledge. Rewriting such prompts to make spatial relationships, object attributes, or stylistic constraints explicit can substantially improve generation quality \cite{meng2024phybench}. This implies that while T2I models are capable of mapping clear textual descriptions to visual outputs, they often struggle with compositional reasoning, object interactions, and fine-grained semantic alignment. These failures highlight a fundamental challenge: \textit{the generation process is sensitive to how explicitly the prompt encodes the intended scene}, and current models lack mechanisms to infer or correct missing semantics during inference.

\section{Test-time Image Refinement (\method)}

To address the limitations of static prompting, we propose \method, a test-time refinement framework that leverages Multi-modal Large Language Models (MLLMs) to iteratively assess and improve prompt-image alignment. Rather than conditioning on a fixed input, \method\ dynamically refines the prompt based on structured feedback from the MLLM, enabling progressive alignment across multiple iterations.

The core of \method\ is a closed-loop refinement process that alternates between \textit{Consistency Analysis} and \textit{Prompt Refinement}. At each step, the MLLM evaluates the semantic consistency between the generated image and the intended prompt, identifying reasoning failures such as incorrect object counts, spatial misplacements, or missing attributes. Using this feedback, the MLLM rewrites the prompt to make implicit semantics explicit, resolve ambiguities, and preserve previously correct elements. The refinement is mainly conditioned on the original prompt, the current image, and the history of prior refinements, enabling context-aware corrections that accumulate over iterations. The complete pipeline is presented in Algorithm~\ref{alg:TDR}.

\begin{figure}
    \centering
    \includegraphics[width=\linewidth]{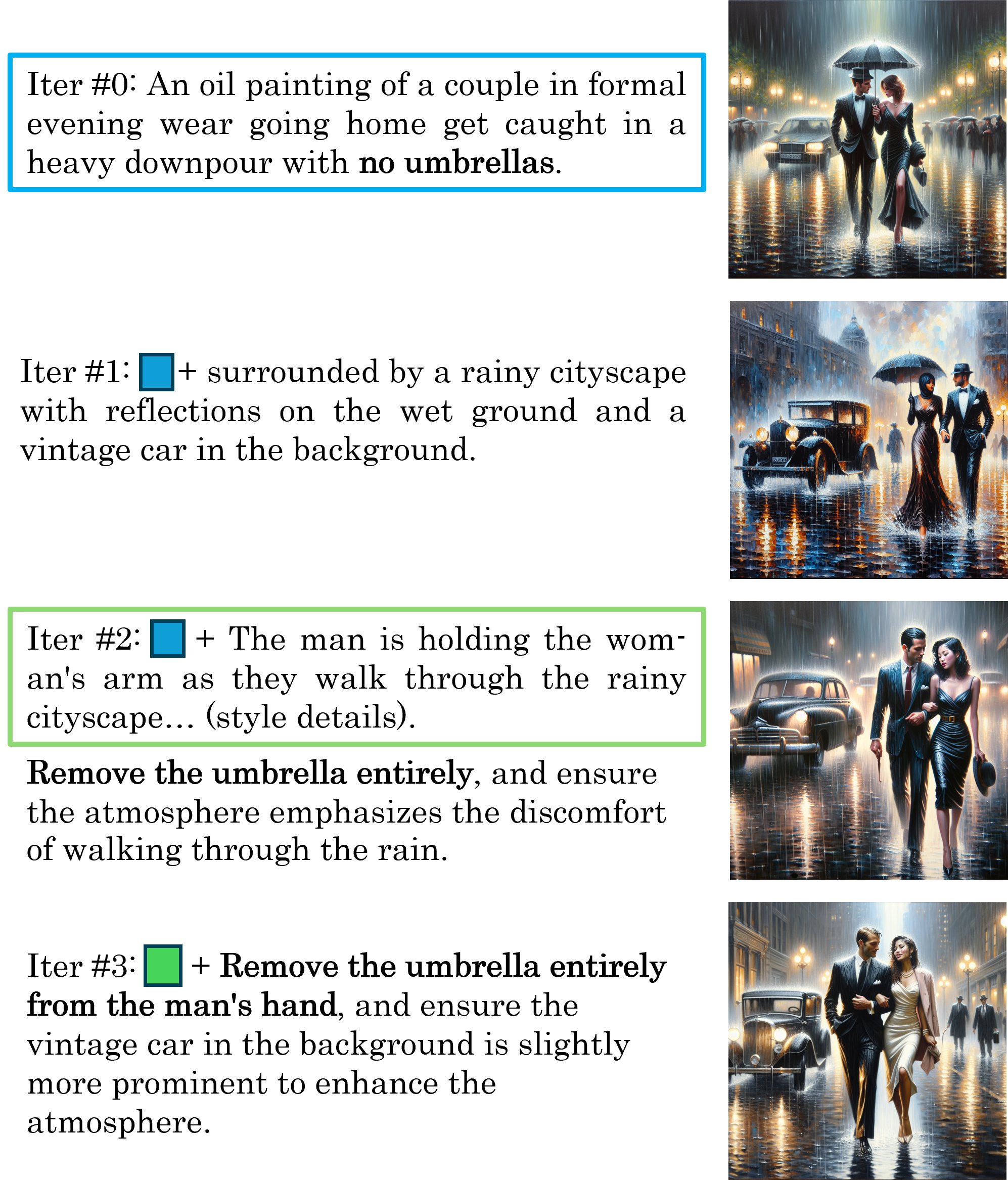}
    \caption{\textbf{Illustrative example of \method.} Each row represents a different iteration of our approach. The first row shows the initial input prompt alongside its generated image. In subsequent rows, the left column presents progressively refined prompts, while the right column shows the corresponding images generated from these refined prompts. Notice how the condition of the umbrella evolves with each refinement. For instance, in the third image, the man holds a small umbrella artifact, which is explicitly addressed in the final image. The colored boxes denote the prompt in the previous iteration. Best viewed in color.}
    \label{fig:example}
\vspace{-0.3cm}
    
\end{figure}

\subsection{Iterative Prompt-Image Alignment via MLLMs}


We begin with a T2I model, parameterized by $\theta$, which generates an image $\mathbf{x}$ given a text prompt $\mathbf{c}$, expressed as:  
\vspace{-0.1cm}
\begin{equation}
    \mathbf{x} = \mathcal{G}_\theta(\mathbf{c}),
\end{equation}  

where $\mathcal{G}_\theta$ represents the generative model responsible for producing the image. The first generation step is performed using the original user-provided prompt $\mathbf{c}_0$, producing an initial image $\mathbf{x}_0$:  
\vspace{-0.1cm}
\begin{equation}
    \mathbf{x}_0 = \mathcal{G}_\theta(\mathbf{c}_0).
\end{equation}  

This initial output serves as the starting point for an iterative refinement process. We define a function $\mathcal{F}_{\text{MLLM}}$, which at iteration $i$, takes as input the sequence of prior prompts $\{\mathbf{c}_{i-1}, \dots, \mathbf{c}_0\}$, the corresponding feedback history $\{\mathbf{f}_{i-1}, \dots, \mathbf{f}_1\}$, and the last generated image $\mathbf{x}_{i-1}$. The refinement process is formulated as:
\vspace{-0.2cm}
\begin{equation}
    \mathbf{c}_{i}, \mathbf{f}_{i} = \mathcal{F}_{\text{MLLM}}(\mathbf{x}_{i-1}, \{\mathbf{c}_{i-1}, \dots, \mathbf{c}_0\}, \{\mathbf{f}_{i-1}, \dots, \mathbf{f}_1\}).
\end{equation}

Here, $\mathcal{F}_{\text{MLLM}}$ first performs prompt-image consistency analysis by evaluating the semantic and structural correctness of $\mathbf{x}_{i-1}$ with respect to the last prompt $\mathbf{c}_{i-1}$, prior refinements, and the accumulated feedback. It identifies issues such as missing objects, incorrect spatial arrangements, attribute mismatches, etc. Based on this feedback \(\mathbf{f}_{i}\), it generates a refined prompt $\mathbf{c}_{i}$ that corrects errors, improves specificity and enhances compositional alignment while preserving fidelity to the original user intent.

The T2I model then regenerates a new image using the updated prompt:  
\begin{equation}
    \mathbf{x}_{i} = \mathcal{G}_\theta(\mathbf{c}_{i}).
\end{equation}  
By iteratively refining the prompt using both the refinement history and the last generated image, the MLLM progressively enforces semantic constraints, making the prompt more explicit and reducing ambiguity. The refinement history serves two purposes: it ensures continuity by retaining valid information from previous iterations, and it helps avoid oscillatory or conflicting corrections. Meanwhile, the last generated image provides direct visual feedback, enabling the MLLM to reason about persistent generation errors and dynamically adjust the prompt based on how the T2I model has interpreted earlier refinements. This history- and context-aware mechanism supports convergence toward semantically accurate and visually grounded generations. Fig.~\ref{fig:method} provides a visual representation of \method.

Since T2I models are highly sensitive to the conditioning prompt, progressively refining the prompt helps reduce semantic ambiguity, mitigate mode collapse, and improve adherence to object attributes, spatial arrangements, and fine-grained details. Consequently, the generated image iteratively improves in visual fidelity and alignment with the user’s intent, while integrating refinements accumulated over previous iterations, as shown in Fig.~\ref{fig:example}. 


The full iterative refinement process is described in \textbf{Algorithm}~\ref{alg:TDR}.





\begin{algorithm}[H]
    \caption{Test-time Image Refinement (TIR)}
    \label{alg:TDR}
    \footnotesize
    \begin{algorithmic}[1]
        \Require $\mathbf{c}_0$, T2I model $\mathcal{G}_\theta$, MLLM refinement function $\mathcal{F}_{\text{MLLM}}$, max iterations $K$
        \State Generate initial image: $\mathbf{x}_0 = \mathcal{G}_\theta(\mathbf{c}_0)$
        \State Initialize prompt history: $\mathcal{C} \gets [\mathbf{c}_0]$
        \State Initialize feedback history: $\mathcal{F} \gets [\ ]$
        \For{$i = 1$ to $K$}
            \State $\mathbf{c}_i, \mathbf{f}_i = \mathcal{F}_{\text{MLLM}}(\mathbf{x}_{i-1}, \mathcal{C}, \mathcal{F})$
            \State Append $\mathbf{c}_i$ to $\mathcal{C}$, $\mathbf{f}_i$ to $\mathcal{F}$
            \State Generate image: $\mathbf{x}_i = \mathcal{G}_\theta(\mathbf{c}_i)$
        \EndFor
        \State \Return $\mathbf{x}_K$ \Comment{Final generated image}
    \end{algorithmic}
\end{algorithm}

\section{Experimental Setup}
\label{sec:experimental_setup}

\subsection{Datasets and Evaluation Metrics}
\label{sec:datasets}
To evaluate Test-time Image Refinement (\method), we use three benchmark datasets, assessing compositional accuracy, prompt comprehension, and complex prompts generalization. Each dataset uses predefined evaluation metrics.


\vspace{-0.2cm}

\paragraph{GENEVAL~\cite{ghosh2023geneval} (Compositional Accuracy)}  
GENEVAL consists of 553 prompts designed to assess fine-grained compositional correctness in text-to-image models. These include object presence, numerical accuracy, spatial relations, and attribute alignment. We evaluate performance using the following:
(i) \textit{Object presence and counting}, using Mask2Former~\cite{cheng2022masked} with a confidence threshold of 0.9 to minimize false positives;
(ii) \textit{Spatial correctness}, determined by comparing centroid positions of detected objects;
(iii) \textit{Attribute correctness}, assessed via zero-shot classification with CLIP ViT-L/14~\cite{radford2021learning}. Final scores are aggregated across tasks and prompts.

\vspace{-0.2cm}

\paragraph{LLM-Grounded Benchmark (Prompt Comprehension)}  
Following~\cite{lian2023llm}, we generate 320 structured prompts covering four key aspects:  
1) Negation (to evaluate whether absent objects remain absent),  
2) Numerical reasoning (to evaluate correct object counts),  
3) Attribute binding (to evaluate correct object-attribute assignments),  
4) Spatial reasoning (to evaluate correct object positioning).  
Our evaluation tests whether TIR improves model adherence to these constraints. We use OWL-ViT~\cite{minderer2022simple} for object detection, bounding box analysis for spatial evaluation, and compute accuracy for each reasoning task.

\vspace{-0.2cm}

\paragraph{DrawBench~\cite{BenchmarkingDiffusion2023} (Generalization to Open-Ended Prompts)} 

DrawBench includes prompts with rare attributes, ambiguous phrasing, and complex styles to test generalization to open-ended generation. Evaluation is based on perceptual quality and prompt adherence. We report scores using \textit{ImageReward}~\cite{xu2023imagereward}, a learned metric based on human preferences, and conduct human evaluation to assess prompt-image consistency.

\subsection{Implementation Details}
We evaluated our method on three types of T2I models (a) Diffusion Models - Stable-Diffusion-1.5 and 2.1 \citep{rombach2022high}, (b) Rectified-Flow Transformer - Flux.1-dev~\citep{flux2024}, and (c) closed-source APIs - DALL-E 3~\citep{dalle3}. Throughout the paper, we kept the denoising steps 100 for diffusion models and 50 for rectified-flow models. We maintained default hyper-parameters for guidance and other model-specific settings. All the models were queried with the same generator seed for reproducibility. For MLLMs, we used GPT-4o as the consistency checker and prompt refiner until otherwise mentioned. The MLLM instruction prompt is shown in Fig.~\ref{fig:refiner-prompt}.

\begin{figure}[htbp]
    \centering
    \includegraphics[width=0.8\linewidth]{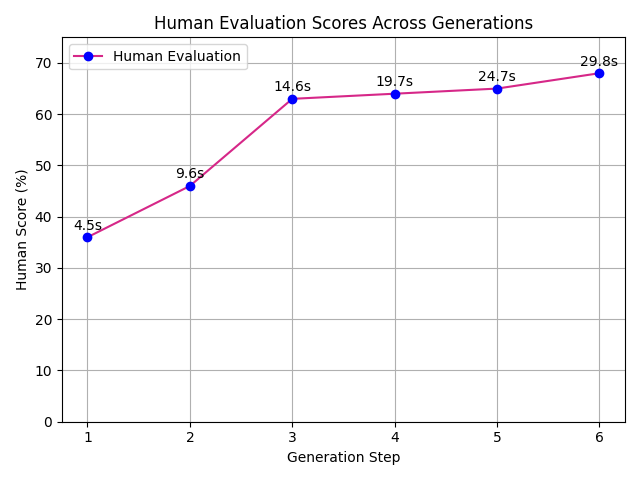}
    \caption{Human evaluation on DrawBench for $K=1$ to $6$ shows $K=3$ balances quality and cost. Best viewed in zoom.}
    \label{fig:tradeoff_human_eval}
    \vspace{-0.3cm}
\end{figure}

After initial generation, the cycle of \textbf{generation $\rightarrow$ assessment $\rightarrow$ refinement $\rightarrow$ regeneration} is repeated for a fixed number of $K=3$ iterations. Empirical analysis showed three iterations provide the best balance between refinement effectiveness and computational cost (see Fig.~\ref{fig:tradeoff_human_eval}). We evaluated 200 DrawBench prompts with two human annotators (Cohen’s \(K = 0.78\)), scoring each image as 1 for perfect alignment with the prompt, and 0 otherwise. Increasing refinements beyond $K=3$ resulted in only marginal improvements while increasing inference time.

\begin{figure}[!ht]

\begin{mdframed}[style=prompt]

\scriptsize
\textbf{You are an Image Improvement Assistant.} Your job is to help make the image more aligned with the \textbf{ORIGINAL} prompt.

\subsection*{Given Inputs:}  
\begin{enumerate}
    \item \textbf{Original User Prompt:}  
    \begin{itemize}
        \item \texttt{\{original\_prompt\}}
    \end{itemize}


    \item \textbf{History of Prompt Refinements and Feedback:}  
    \begin{itemize}
        \item \texttt{\{history\_text\}}
    \end{itemize}

    \item \textbf{Current Image Analysis:}
    \begin{itemize}
        \item Look at the image and identify what aspects DIFFER from what the \textbf{ORIGINAL} prompt requested.
        \item Analyze what essential elements from the \textbf{ORIGINAL} prompt are missing or incorrectly represented.
    \end{itemize}
\end{enumerate}

\subsection*{Your Task:}  
\begin{enumerate}
    \item Create a \textbf{NEW PROMPT} that will help generate an image that better matches the \textbf{ORIGINAL} prompt.
    \item Focus on fixing what's missing or incorrectly represented in the current image
    \item Incorporate suitable elements from \textbf{prompt history} into the \textbf{NEW PROMPT}.
    \item The goal is to get progressively closer to fulfilling the \textbf{ORIGINAL} prompt.
\end{enumerate}

\subsection*{Output:}  
\textbf{REFINED PROMPT:}  
\begin{center}
    \texttt{"<A detailed and enhanced version of the last prompt that improves alignment>"}
\end{center}

\end{mdframed}

\caption{Refinement prompt provided to the MLLM.}
\label{fig:refiner-prompt}

\end{figure}

\vspace{-0.2cm}
\section{Results}

\begin{figure*}[t]
    \centering
    \hspace{3cm}\includegraphics[width=0.9\linewidth]{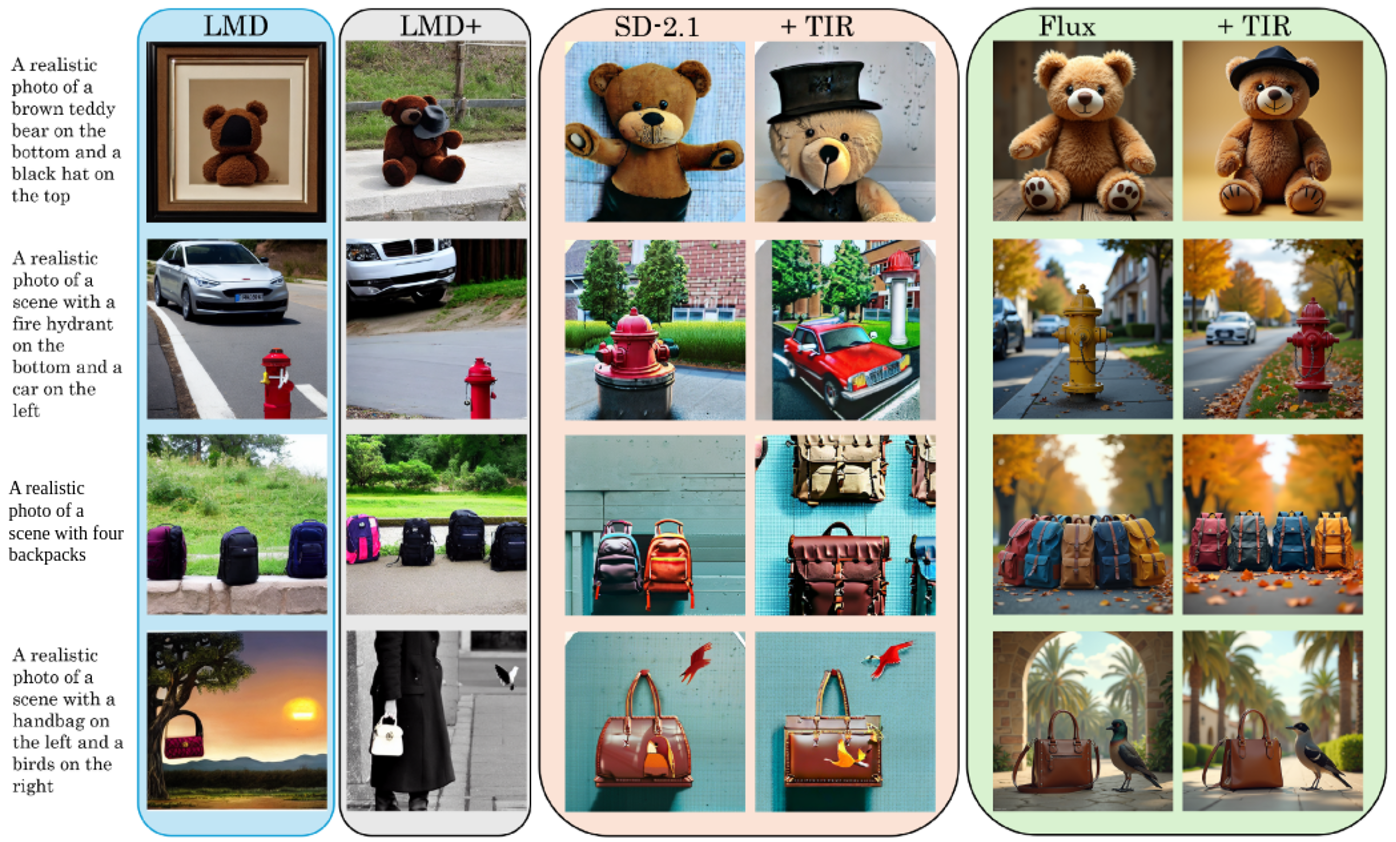}
    \caption{\textbf{Qualitative comparison of \methodwithspace against the LMD and LMD+ baselines on prompts from the LLM‐Grounded Diffusion Benchmark.} Each column corresponds to one method’s outputs for the same input prompt. Although LMD and LMD+ are tuned for this benchmark, \methodwithspace consistently produces more accurate images aligned with the given prompts. In the final row, where the baseline already meets the prompt requirements, \methodwithspace preserves correctness without introducing any regression.}
    \label{fig:qualitative-analysis}
\end{figure*}

\begin{figure*}[!ht]
    \centering
    \hspace{0cm}\includegraphics[width=0.8\linewidth]{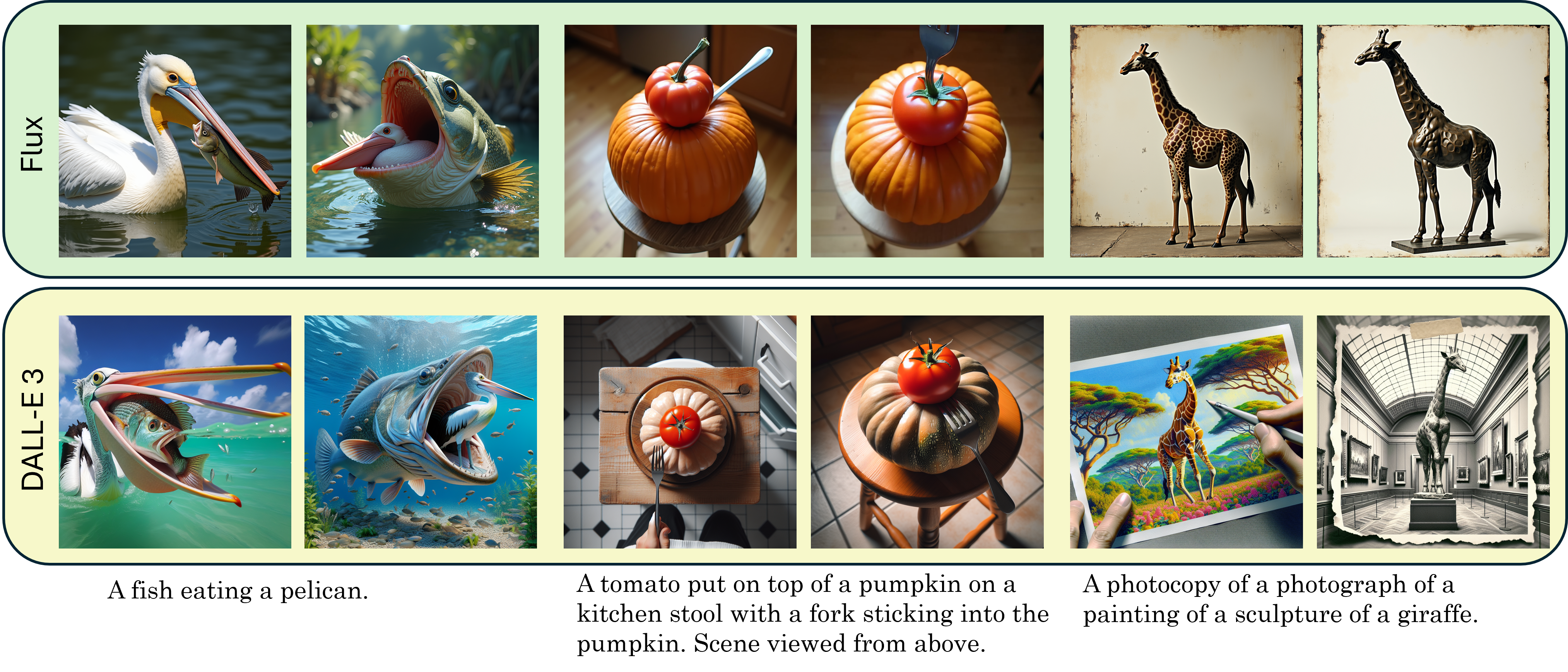}
    \caption{\textbf{Qualitative results of \methodwithspace on open-ended prompts from DrawBench.} Images on the left are from base model while the \textbf{right image is after \methodwithspace refinement}. Even stronger T2I models struggle to disentangle semantic components in complex prompts, \methodwithspace extracts capabilities of these models by refining prompts into simpler sections while retaining the original instruction, giving desired outputs.}
    \label{fig:qualitative-analysis2}
\vspace{-0.3cm}
    
\end{figure*}


\begin{table*}[t]
\centering
\footnotesize
\begin{tabular}{lccccc}
    \toprule
    \textbf{Method} & \textbf{Negation} & \textbf{Numeracy} & \textbf{Attribute} & \textbf{Spatial} & \textbf{Average} \\
    \midrule
    MultiDiffusion~\cite{bar2023multidiffusion}     & 29.0 & 28.0 & 26.0 & 39.0 & 30.5 \\
    Backward Guidance~\cite{chen2024training}    & 22.0 & 37.0 & 26.0 & 67.0 & 38.0 \\
    BoxDiff~\cite{xie2023boxdiff}              & 22.0 & 30.0 & 37.0 & 71.0 & 40.0 \\
    {\color{gray} LayoutGPT + GLIGEN}~\cite{feng2023layoutgpt}   & {\color{gray} 36.0} & {\color{gray} 65.0} & {\color{gray} 26.0} & {\color{gray} 78.0} & {\color{gray} 51.3} \\
    LMD~\cite{lian2023llm} & 100.0 & 40.0 & 50.5 & 38.7 & 57.3 \\
    {\color{gray} LMD+~\cite{lian2023llm}}           & {\color{gray} 100.0} & {\color{gray} 50.9} & {\color{gray} 63.5} & {\color{gray} 42.0} & {\color{gray} 64.1} \\
    \midrule
    SD-1.5~\cite{rombach2022high}         & 35.0  & 40.0 & 42.0 & 38.0 & 38.75 \\
    \quad w/ \method     & \cellcolor{green!10}80.0 & \cellcolor{green!10}51.7 & 30.0 & \cellcolor{green!10}54.5 & \cellcolor{green!10}54.0 (\textbf{\textcolor{darkgreen}{+15.25}}) \\
    SD-2.1~\cite{rombach2022high}         & 45.0  & 50.6 & 18.0 & 44.5 & 39.5 \\
    \quad w/ \method     & \cellcolor{green!10}85.0 & \cellcolor{green!10}56.0 & \cellcolor{green!10}18.0 & 43.0 & \cellcolor{green!10}50.5 (\textbf{\textcolor{darkgreen}{+11.0}}) \\
    Flux~\cite{flux2024}           & 10.0  & 52.2 & 83.0 & 81.5 & 56.7 \\
    \quad w/ \method     & \cellcolor{green!10}55.0 & \cellcolor{green!10}64.7 & 81.8 & 77.0 & \cellcolor{green!10}\textbf{69.6} (\textbf{\textcolor{darkgreen}{+12.9}}) \\
    DALL-E 3~\cite{dalle3}        & 31.6     & 44.5   & 73.0   & 81.0   & 57.5   \\
    \quad w/ \method     & \cellcolor{green!10}73.7    & \cellcolor{green!10}49.2   & 71.0   & \cellcolor{green!10}83.0   & \cellcolor{green!10}69.2 (\textbf{\textcolor{darkgreen}{+11.7}})   \\
    \bottomrule
    \end{tabular}%
    
    \vspace{0.4mm}
    \caption{\textbf{Comparison of training-free image tuning methods on LLM-Grounded Diffusion Benchmark.} We observe consistent improvements with \methodwithspace applied on various T2I models. Further, the gains are more substantial in spatial and numeracy tasks, compared to color attribution which is consistent to our observation on GENEVAL benchmark. Note that, \textcolor{gray}{LMD+}~\cite{lian2023llm} and \textcolor{gray}{GLIGEN}~\cite{feng2023layoutgpt} use a specialized bbox (layout) tuned diffusion model.}
\vspace{-0.2cm}
    
    \label{tab:llm-grounded-diffusion-benchmark}
\end{table*}

\begin{table*}[t]
\centering
\footnotesize
\vspace{-2mm}
\begin{tabular}{l|c|c|c|c|c|c|c}
\toprule
Model      & Position & Counting & Single Obj. & Two Object & Color Attr & Colors & Overall Score\\
\toprule
Flux~\cite{flux2024} & 19.00    & 68.75    & 100.00        & 75.76      & 48.00      & 77.66  & 64.86\\
\quad w/ \method & \cellcolor{green!10}49.00  & \cellcolor{green!10}71.25  & 98.75       & \cellcolor{green!10}80.81  & 47.00      & \cellcolor{green!10}80.85  & \cellcolor{green!10}71.27 (\textbf{\textcolor{darkgreen}{+6.41}})\\
\midrule
DALL-E 3~\cite{dalle3}     & 34.00    & 48.75    & 96.25         & 77.78      & 31.00      & 74.47  & 60.37\\
\quad w/ \method& \cellcolor{green!10}45.00  & \cellcolor{green!10}60.00  & \cellcolor{green!10}96.25       & \cellcolor{green!10}82.83  & \cellcolor{green!10}38.00  & \cellcolor{green!10}86.17  & \cellcolor{green!10}68.04 (\textbf{\textcolor{darkgreen}{+7.67}})\\
\hline
\end{tabular}

\caption{\textbf{Performance on GENEVAL benchmark.} Here we present accuracies across six fine-grained categories, along with overall scores. Across all categories, integrating \methodwithspace leads to the largest improvement in  spatial positioning and counting tasks, while single‐object prompts remain highly accurate even without \method. In more complex cases requiring multiple objects and  color understanding, \methodwithspace delivers consistent gains for stronger T2I models. Overall, all T2I models experience a substantial increase in their overall scores.}
\vspace{-0.2cm}

\label{tab:geneval-benchmark}
\end{table*}

\paragraph{Qualitative results.}  
Fig.~\ref{fig:qualitative-analysis} presents a visual comparison against baselines on LLM-grounded diffusion benchmark prompts. In general, base models struggle to capture the compositional nature of the prompt. Baselines~\citep{lian2023llm,qu2023layoutllm} fail to maintain image quality due to their reliance on intermediate noisy latent pasting, a limitation inherent to diffusion-based methods. \methodwithspace enhances T2I model capabilities by breaking down complex compositional attributes into structured semantic components, allowing the model to better capture fine-grained details and spatial relationships. As a result, TIR generates images with higher fidelity, improved clarity, and stronger alignment with the given prompt.

In Fig.~\ref{fig:qualitative-analysis2}, we showcase the capabilities of \methodwithspace on more powerful models. As the prompt becomes more structured and semantically meaningful, a model with a robust language backbone is able to more effectively leverage the decomposed semantic components, thereby synthesizing images that not only exhibit enhanced visual quality but also preserve intricate compositional details. This synergy between the refined semantic understanding and image generation leads to outputs that are both visually compelling and semantically coherent, further highlighting the efficacy and scalability of our approach. More qualitative images are provided in the supplementary material, showing full trajectories of prompt refinement and generation. 


\vspace{-0.2cm}

\paragraph{Comparison of training-free image tuning methods on LLM-Grounded Diffusion Benchmark.} In Table~\ref{tab:llm-grounded-diffusion-benchmark}, we demonstrate that our method improves control on image generation across different models. We verify that \methodwithspace works on different types of image generation models, irrespective of their (a) backbone architecture - U-Net ~\citep{rombach2022high} or DiT~\cite{dalle3}; (b) generative modeling - diffusion~\citep{rombach2022high} or rectified-flow based~\citep{flux2024}; and (c) access type - open-sourced~\citep{rombach2022high,flux2024} or close-sourced~\citep{dalle3}. Further, we have observed the largest gains of \methodwithspace in the Negation and Numeracy tasks. We attribute this distinction to the ability of the base image generation model and their ability to understand detailed prompts. As models evolve, their ability to process semantically equivalent yet more detailed prompts improves, as reflected in the higher average scores achieved with \methodwithspace by Flux and DALL-E 3 compared to previous-generation Stable Diffusion models.
 Finally, we observe that \methodwithspace outperforms training-free baselines~\cite{bar2023multidiffusion, chen2024training, xie2023boxdiff, lian2023llm}. Although LMD+ achieves high accuracy, it is not a direct comparison, as it uses GLIGEN~\citep{li2023gligen}, which is a fine-tuned diffusion model to take grounded bounding boxes as input.


\begin{figure}
    \centering
    \includegraphics[width=0.825\linewidth]{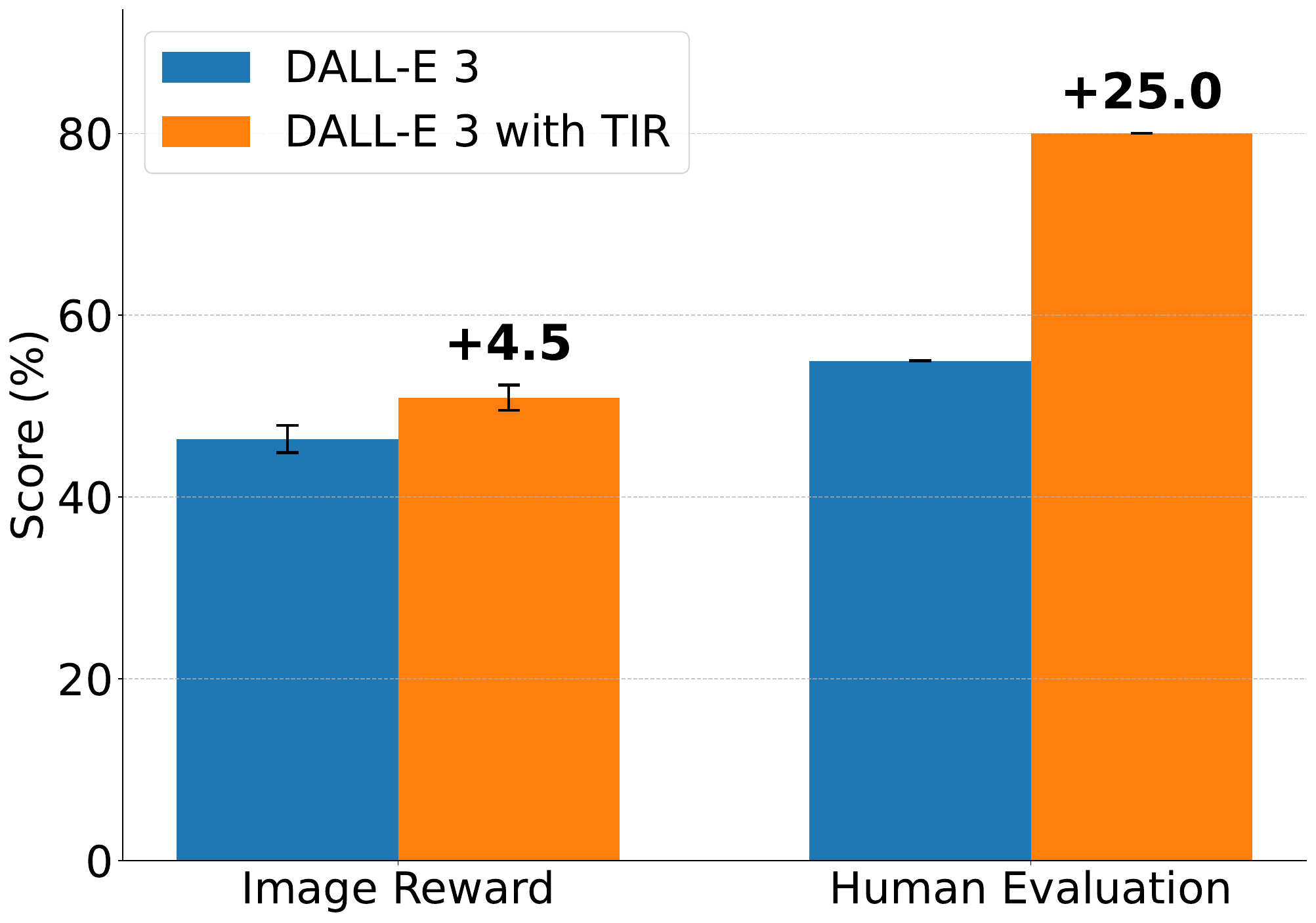}
    \caption{\textbf{Evaluation on Drawbench dataset.} We compare normalized \textit{ImageReward} \citep{xu2023imagereward} score on DALL-E 3 model and observe consistent improvement over multiple tries. The error bars show the \textit{std dev.} over five runs. Further, we also performed human evaluation for prompt-image correctness on Drawbench due to the open-ended nature of its prompt and display substantial gains.}
    \label{fig:drawbench-graph}
\end{figure}

\vspace{-0.3cm}

\paragraph{Comparison on GENEVAL benchmark.} We evaluate fine-grained compositional control achieved by \methodwithspace in Table~\ref{tab:geneval-benchmark}. We observe consistent improvements with \methodwithspace in both Flux.1-dev \cite{flux2024} and DALL-E 3 \cite{dalle3} models across all categories. These results corroborate our finding that the performance of \methodwithspace is enhanced by stronger base diffusion models. In addition, we show an increase in quantitative performance in all composition categories, showcasing the strength of \methodwithspace in image generation.

\begin{table*}[t]
\centering
\footnotesize
\vspace{-2mm}
\begin{tabular}{l|c|c|c|c|c|c|c}
\toprule
Model      & Position & Counting & Single Obj. & Two Object & Color Attr & Colors & Overall Score\\
\toprule
Flux~\cite{flux2024} & 19.00    & 68.75    & 100.00        & 75.76      & 48.00      & 77.66  & 64.86\\
\quad w/ \method ( Qwen-$2.5$-$7$B) & \cellcolor{green!10}29.00  & 67.50 & 98.75       & \cellcolor{green!10}86.87  & 44.00      & \cellcolor{green!10}81.91  & \cellcolor{green!10}68.01 (\textbf{\textcolor{darkgreen}{+3.15}}) \\
\bottomrule
\end{tabular}
\caption{\textbf{Comparison against different MLLMs.} We evaluated GENEVAL benchmark with Qwen-$2.5$-$7$B as prompt-image checker. Coupled with Flux results on GPT-4o from Tab.~\ref{tab:geneval-benchmark}, we show steady gains with another MLLM. Interestingly, we observe similar trends between the two MLLMs, indicating certain biases in the base T2I model.}
\label{tab:geneval-qwen}
\vspace{-0.2cm}

\end{table*}

\vspace{-0.5cm}

\paragraph{Evaluation on Drawbench.} Evaluating image quality in a compositional setting is a challenging task. Following~\citep{ma2025inference}, we evaluated \emph{ImageReward} metrics on DrawBench against DALL-E 3 model with GPT-4o as prompt refiner. \emph{ImageReward} possess more nuanced evaluative aspects, capturing \emph{Aesthetic Score} and \emph{CLIPScore} and closely align with human preferences, making it suitable for image quality evaluation metric~\citep{ma2025inference}. Fig.~\ref{fig:drawbench-graph} shows the increase in image quality with \method. To show consistent gains, we sample five images for each prompt and report the mean and standard deviation in the score.
We also performed human-evaluation to verify prompt-image correctness on Drawbench. Human annotators gave a score of $1$ if the image is perfectly aligned to the prompt and $0$ otherwise. Fig.~\ref{fig:drawbench-graph} highlights \methodwithspace results in prompt-image correctness. We report a $25$\% higher alignment towards the prompt compared to the vanilla model and a $4.5$\% gain in \emph{ImageReward} score.




\begin{figure}
    \centering
    \includegraphics[width=0.9\linewidth]{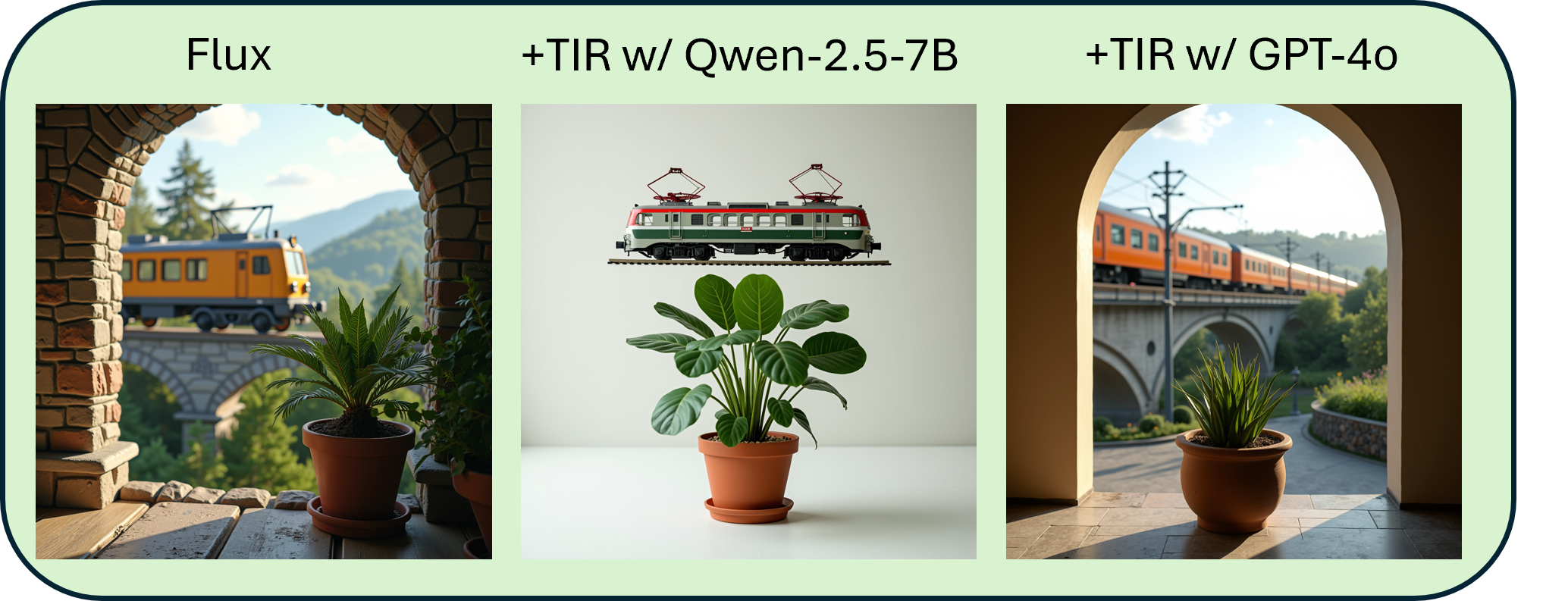}
    \caption{\textbf{Qualitative comparison with different MLLMs.} Input prompt - \textit{a photo of a train above a potted plant}. While Qwen-$2.5$-$7$B produces a semantically correct output, GPT-4o optimizes the prompt to produce a more natural image.}
    \label{fig:ablation}
    \vspace{-0.3cm}

\end{figure}

\vspace{-0.2cm}

\paragraph{Comparison against different MLLMs.} We evaluate the effectiveness of TIR refinement across different MLLMs. Specifically, we use the Qwen-$2.5$-$7$B MLLM~\citep{qwen2.5, qwen2} and test it on the GENEVAL benchmark~\cite{ghosh2023geneval} with Flux.1-dev~\citep{flux2024} as the image generator. Table~\ref{tab:geneval-qwen} demonstrates consistent improvements with \methodwithspace when using Qwen-$2.5$-$7$B as the refinement MLLM. It highlights the versatility of \methodwithspace and its wider applicability with different MLLMs. In Fig.~\ref{fig:ablation}, we show visually how different MLLMs result in stark outcomes. While Qwen-$2.5$-$7$B provided a semantically correct prompt for the input - \textit{a photo of a train above a potted plant}, GPT-4o refined the prompt to produce a more natural image. Interestingly, comparing with Flux + GPT-4o performance reported in Table~\ref{tab:geneval-benchmark} aligns with our intuition that stronger MLLMs refine complex prompts more easily, resulting in better control in generated images. Further, \emph{Color Attr.} does not experience improvement with either MLLMs which suggests that the base T2I model (Flux) has a limitation in identifying correct color attributions. Perhaps, \methodwithspace could be a way of identifying bias in text-to-image models. We leave this exploration for future works.


\vspace{-0.2cm}
\section{Limitations}

\begin{figure}[htbp]
    \centering
    \includegraphics[width=\linewidth]{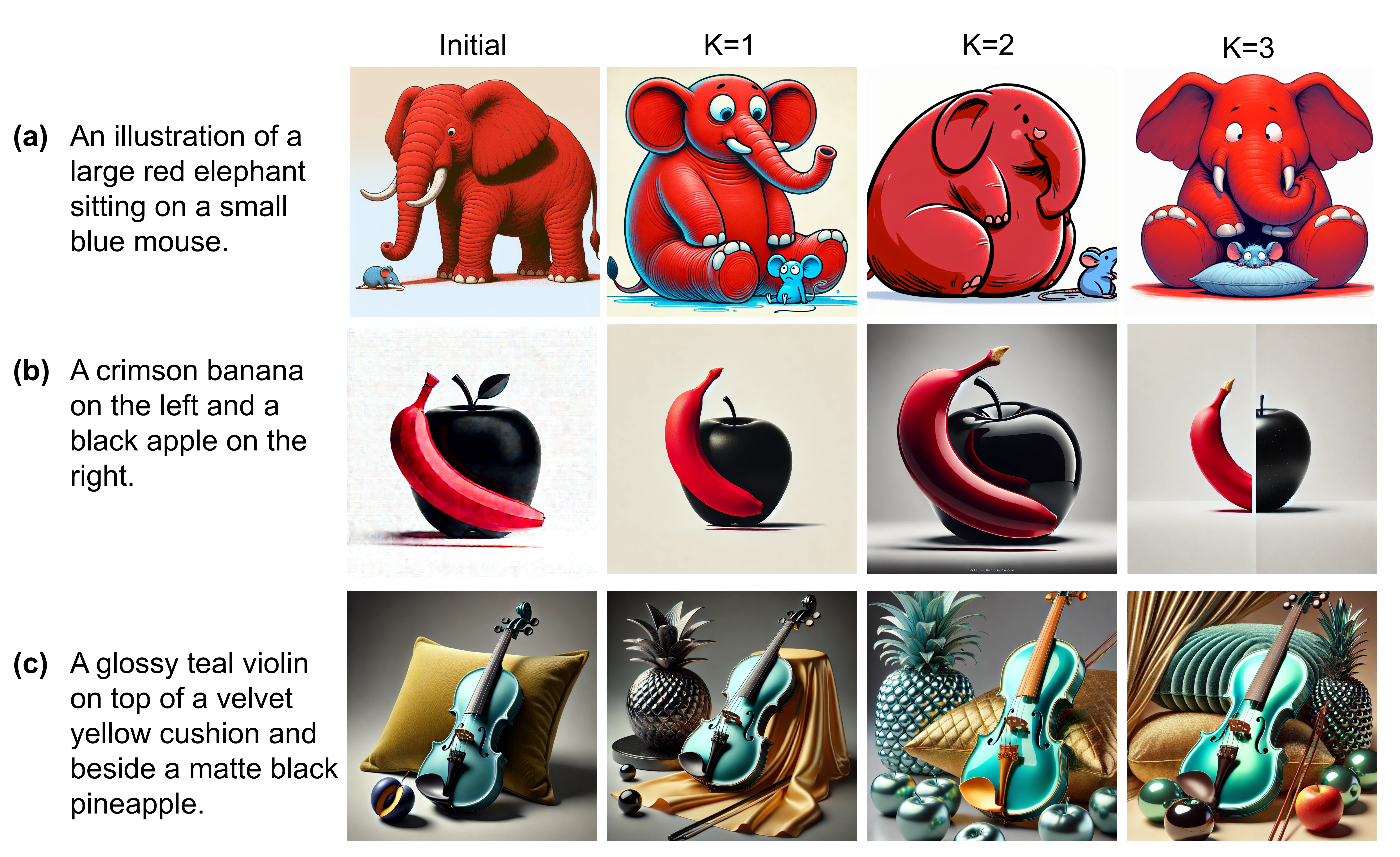}
    \caption{Failure cases using  DALL-E 3 as the T2I model and Qwen-$2.5$-$7$B as the MLLM. \textbf{(a):} Demonstrates a \textit{semantic misalignment} between the intended object interaction and the generated outputs. Across iterations, the T2I model fails to depict the red elephant physically sitting on the blue mouse.  
    \textbf{(b):} Reveals a \textit{vocabulary--distribution mismatch}, where the crimson banana and black apple remain consistent in form but the T2I model struggles to enforce the intended left-right layout, resulting instead in a split view composition in the final image.  
    \textbf{(c):} Shows an \textit{attribute regression}, where a matte black pineapple shifts to teal in later iterations and the unintended teal tone also appears in surrounding elements, indicating overfitting to dominant visual styles.}
    \label{fig:tradeoff}
    \vspace{-0.3cm}
\end{figure}

Although TIR is model agnostic and applies to all T2I models, it does have certain limitations. Firstly, \methodwithspace relies on the strong language backbone of a T2I model. T2I Models with limited semantic understanding or poor comprehension of complex prompts may show minimal improvement. In contrast, a stronger language backbone observes substantial prompt-image alignment as observed with DALL-E 3. We believe that with the evolution of T2I models the applicability of \methodwithspace would only increase. Secondly, it is possible that there is a misalignment between the vocabulary of the T2I model and the prompt refiner MLLM. A semantically preserving refined prompt might not bring any change in the output of the T2I model if its vocabulary distribution differs from that of the MLLM (see Fig.~\ref{fig:tradeoff} (a), (b)). However, we believe that with sufficient refinement steps such a gap could be overcome. Moreover, the MLLM itself may sometimes introduce flawed refinements when it incorrectly identifies an issue in a generated image. These errors can persist over iterations, leading to attribute regressions or unintended style dominance, as illustrated in Fig.~\ref{fig:tradeoff} (c). Perhaps, a better refinement strategy can overcome this, and we implore the community to explore this direction.

\vspace{-0.4cm}

\section{Conclusion}
\label{sec:conclusion}  

We addressed the sensitivity of foundational T2I generation models to minor prompt variations, which often results in misaligned and inconsistent image outputs. Since manual prompt modifications are costly and impractical at scale, we introduced a test-time prompt refinement framework that iteratively improves prompts using an MLLM. Our approach enhances alignment and visual coherence without altering the underlying T2I model, leading to both quantitative and qualitative improvements across multiple benchmarks. 

{
    \small
    \bibliographystyle{ieeenat_fullname}
    \bibliography{main}
}

\clearpage
\setcounter{page}{1}
\maketitlesupplementary
\section{Dataset and Evaluation Details}

\subsection{Benchmark Datasets}

We use three benchmark datasets to assess compositional fidelity, prompt comprehension, and generalization:

\subsubsection{GENEVAL: Compositional Accuracy}

GENEVAL [14] consists of 553 prompts testing object presence, count, color accuracy, spatial positioning, and attribute binding. It provides structured evaluation for fine-grained correctness in T2I models.

\subsubsection{LLM-Grounded Diffusion Benchmark: Prompt Comprehension}

Lian et al. [26] designed a benchmark to assess how well T2I models interpret prompts. Based on their given templates, we created 320 structured prompts, using 20 most common COCO objects, covering:
\begin{itemize}
    \item Negation (A realistic photo of a scene without [object name])
    \item Numerical reasoning (A realistic photo of a scene with [number] [object name])
    \item Attribute binding (A realistic photo of a scene with [modifier 1] [object name 1] and [modifier 2] [object name 2])
    \item Spatial reasoning (A realistic photo of a scene with [object name 1] on the [location] and [modifier 2] [object name2] on the [opposite location], where the location is chosen from left, right, top, and bottom.)
\end{itemize}

\subsubsection{DrawBench: Generalization to Open-Ended Prompts}

DrawBench [42] evaluates generative adaptability across ambiguous descriptions, numerical constraints, spatial relations, and rare words. It helps measures perceptual plausibility rather than strict correctness.

\section{Qualitative Results}
\begin{figure*}
    \centering
    \begin{subfigure}[b]{\linewidth}
        \includegraphics[width=\linewidth]{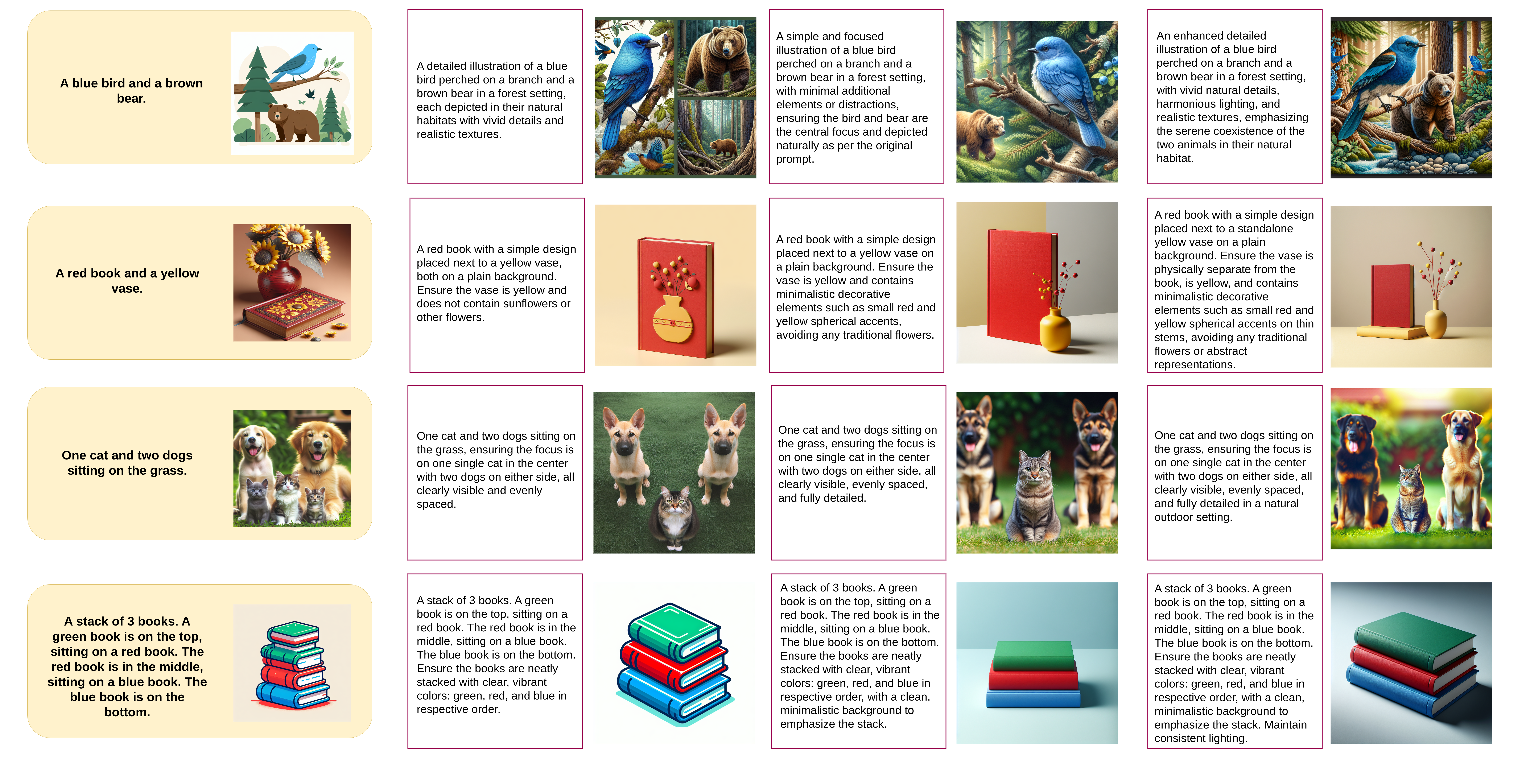}
    \end{subfigure}
    
    \vspace{-0.2cm} 

    \begin{subfigure}[b]{\linewidth}
        \includegraphics[width=\linewidth]{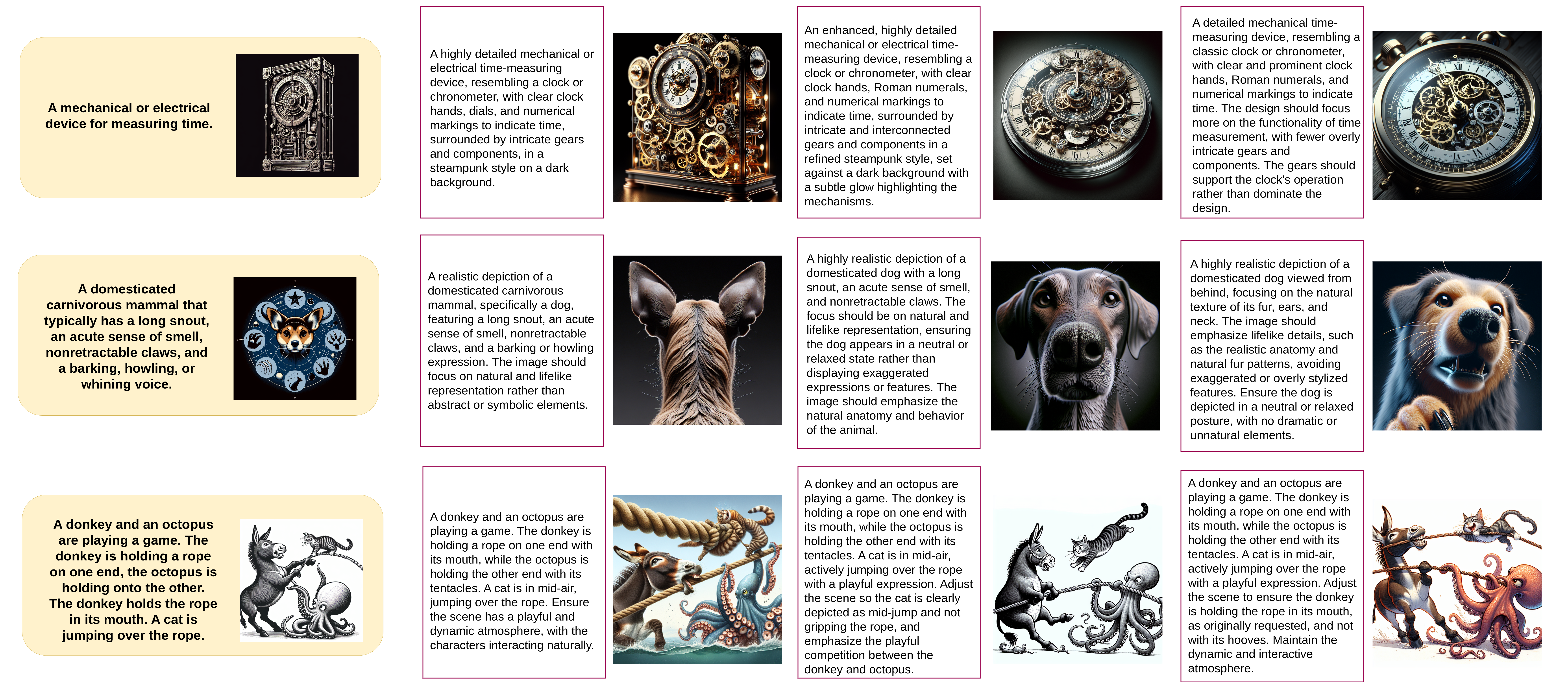}
    \end{subfigure}
    
    \caption{\textbf{Qualitative results on Drawbench using DALL-E 3 as the T2I model and GPT-4o as the MLLM.} Each row shows the prompt refinement trajectory, beginning with the initial prompt and generation on the left, followed by three rounds of GPT-4o-guided refinements. The results demonstrate how TIR progressively enhances alignment between user intent and visual output.}
    \label{fig:drawbench_results1}
\end{figure*}

\begin{figure*}
    \centering
    \begin{subfigure}[b]{\linewidth}
        \includegraphics[width=\linewidth]{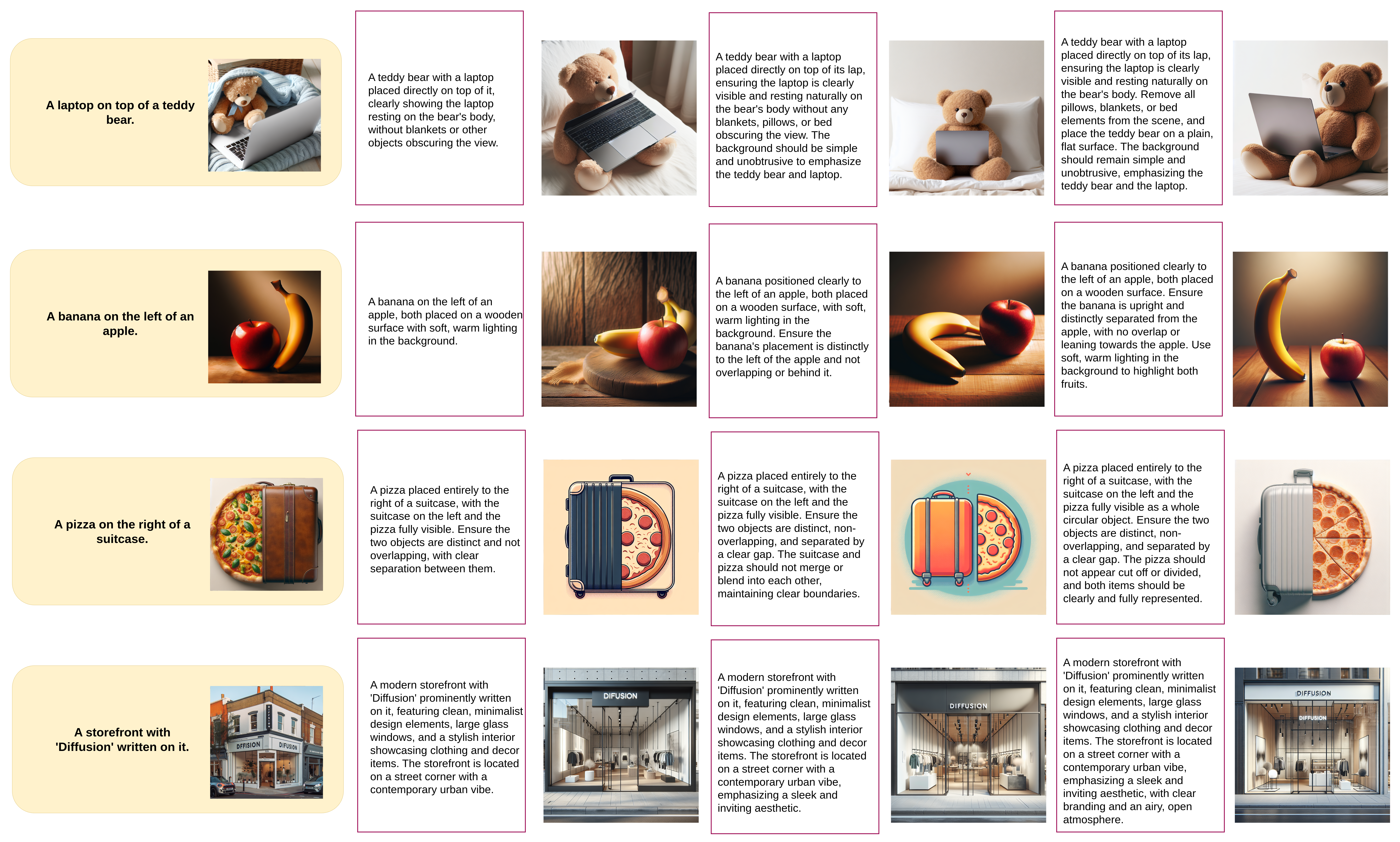}
    \end{subfigure}
    
    \vspace{0.0cm} 

    \begin{subfigure}[b]{\linewidth}
        \includegraphics[width=\linewidth]{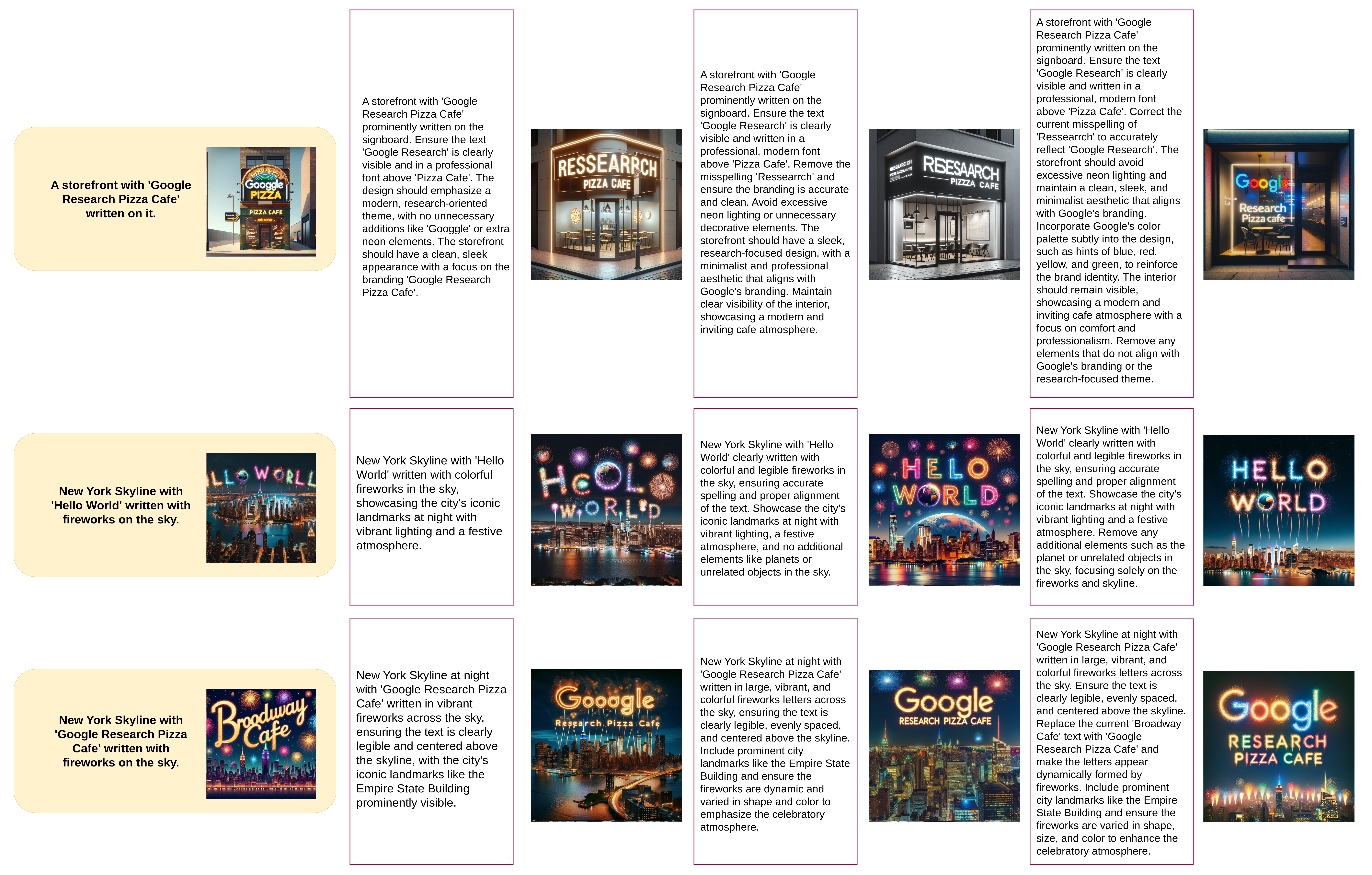}
    \end{subfigure}
    
    \caption{\textbf{Qualitative results on Drawbench using DALL-E 3 as the T2I model and GPT-4o as the MLLM.} Each row shows the prompt refinement trajectory, beginning with the initial prompt and generation on the left, followed by three rounds of GPT-4o-guided refinements. The results demonstrate how TIR progressively enhances alignment between user intent and visual output.}
    \label{fig:drawbench_results2}
\end{figure*}

Figs.~\ref{fig:drawbench_results1} and \ref{fig:drawbench_results2} showcase qualitative results of our Test-time Iterative Refinement (TIR) method on Drawbench prompts, using GPT-4o as the MLLM for prompt refinement and DALL-E 3 as the text-to-image (T2I) generator. Each row illustrates the initial prompt and image on the left, followed by three iterations of refinement. For each iteration, the left column presents the progressively refined prompts, while the right column shows the corresponding images generated from these refined prompts. We observe progressive improvements in visual-semantic alignment, demonstrating TIR’s ability to iteratively correct misalignments and converge toward more faithful generations with respect to original user intent.

\end{document}